# Detecting Causal Relations in the Presence of Unmeasured Variables

**Peter Spirtes**
Department of Philosophy
Carnegie Mellon University

## Abstract

The presence of latent variables can greatly complicate inferences about causal relations between measured variables from statistical data. In many cases, the presence of latent variables makes it impossible to determine for two measured variables A and B, whether A causes B, B causes A, or there is some common cause. In this paper I present several theorems that state conditions under which it is possible to reliably infer the causal relation between two measured variables, regardless of whether latent variables are acting or not.

## 1 Introduction

The problem of inferring causal relations from statistical data in the absence of experiments arises repeatedly in many scientific disciplines, including sociology, economics, epidemiology, and psychology. In addition, the building of expert systems could be expedited if background knowledge elicited from experts could be supplemented with automated techniques. Recently, efficient algorithms for determining causal structure (in the form of Bayesian networks) from statistical data when there are no unmeasured or "latent" variables have been proposed. (See Spirtes, Glymour and Scheines 1990, Spirtes and Glymour 1991, Spirtes, Glymour, and Scheines forthcoming, Verma and Pearl 1990, and Pearl and Verma 1991.)

Inferring causal relations when unmeasured variables are also acting is a much more difficult problem. In many cases it is impossible to infer the structure among the latent variables from statistical relations among the measured variables. But the presence of latent variables can also make it difficult to infer the causal relations among the measured variables themselves. One important question for many policy decisions is "Does A cause B?" Statistics about A and B alone do not suffice to answer these questions. When only two variables, A and B, have been measured, and there is a correlation between the two, this does not suffice to establish whether A caused B, B caused A, or there is a third variable causing both A and B. Nevertheless, when more variables are measured, more knowledge about the causal relations between A and B is possible. Drawing upon recent results of Verma and Pearl (Verma and Pearl 1990, and Pearl and Verma 1991), I will prove in Theorem 2 that there are some circumstances in which it is possible to establish that A caused B, rather than that B caused A, or that a third variable caused both A and B; I will prove in Theorem 3 that there are other circumstances in which the possibility that A caused B can be eliminated. The proofs are given in the Appendix. I will also demonstrate that a recent proposal derived from Pearl and Verma (1991) to establish more general conditions for causal pathways in the presence of unmeasured causes is incorrect.

## 2 Results

Causal processes between a set of random variables **V** are represented by a directed acyclic graph over **V**, where there is an edge from A to B if and only if A is an immediate cause of B relative to **V**. (For a discussion of the meaning of immediate causation, see Spirtes, Glymour and Scheines 1991.) If there is a directed path from A to B in the causal graph, I will say that A is a (possibly indirect) cause of B.

If a distribution is placed over the exogenous variables in a causal process (variables of zero indegree in the causal graph), which in turn affect the values of other random variables, the result is a joint distribution over all of the random variables. In that case, I will say that the causal process generated the joint distribution. Following Pearl (1988) I assume that the distribution generated by a causal process satisfies the Minimality and Markov conditions for the causal graph of that process. In Pearl's terminology (Pearl 1988) the causal graph is a **Bayes network** of any distribution that it generates. (In what follows, I will capitalize random variables, and boldface any sets of variables.)

**Markov Condition:** Let Descendant(V) be the set of descendants of V in a graph G, and Parents(V) be the set of parents of V in G. A graph G and a probability distribution P on the vertices **V** of G satisfy the Markov condition if and only if for every V in **V**, and every subset

X of **V**, V and X\{V} ∪ Descendants(V) are independent conditional on Parents(V).

**Minimality Condition**: A graph G and a probability distribution P satisfies the minimality condition if and only if G and P satisfy the Markov condition and for every graph H obtained by deleting an edge from G, H and P do not satisfy the Markov condition.

If P satisfies the Markov condition for graph G, and every conditional independence true of the distribution P is entailed by the Markov condition, then we say that P is **faithful** to G.

In a directed graph G, I will write X -> Y if there is an edge from X to Y in G. In an undirected graph U, I will write X - Y if there is an undirected edge between X and Y. X and Y are **adjacent in a directed graph** G if and only if either X -> Y or Y -> X in G. Two edges are **adjacent in an undirected graph** U if and only if X - Y in U. In a directed acyclic graph G, an **undirected path U from X to Y** is a sequence of vertices starting with X and ending with Y such that for every pair of variables A and B that are adjacent to each other in the path, A and B are adjacent in G, and no vertex occurs more than once in U. In a directed acyclic graph G, a **directed path P from X to Y** is a sequence of vertices starting with X and ending with Y such that for every pair of variables A and B that are adjacent to each other in the path in that order, the edge A -> B occurs in G, and no vertex occurs more than once in P. An **edge between X and Y occurs in a path P** (directed or undirected) if and only if X and Y are adjacent in P. If an undirected path U contains an edge between X and Y, and an edge between Y and Z, the two edges **collide** at Z if and only if X -> Y and Z -> Y in G. On an undirected path U, Z is an **unshielded collider** if and only if there exist edges X -> Y and Z -> Y in U, and Z and X are not adjacent in G. X is an **ancestor** of Y and Y is a **descendant** of X if and only if there is a directed path from X to Y.

Pearl and Verma have shown how to calculate the conditional independence relations that are entailed by distributions satisfying the Markov condition for a graph G using the d-separability relation. In graph G, variables X and Y are **d-separated** by a set of vertices **S** not containing X or Y if and only if there exists no undirected path U between X and Y, such that (i) every collider on U has a descendent in S and (ii) no other vertex on U is in S. Disjoint sets of variables X and **Y** are d-separated by **S** in G if and only if every member of X is d-separated from every member of **Y** by **S** in G. If distribution P satisfies the Markov condition for graph G, then the Markov condition entails that X is independent of **Y** conditional on **S** if and only if X is d-separated from **Y** by **S** in G (Pearl 1988).

The following algorithm (Spirtes 1990, Spirtes 1991) reconstructs the set of all graphs of causal processes that could have generated a given probability distribution, under the assumptions that no latent variables are present (i.e. every cause of a pair of measured variables is itself measured) and that the distribution is faithful to the graph of the causal process that generated it. The d-separability relations of the graph can be determined either by performing tests of conditional independence on the generated distribution, or in the linear case, tests of zero partial correlations.

Let $A_C(A,B)$ denote the set of vertices adjacent to A or to B in graph C, except for A and B themselves. Let $U_C(A,B)$ denote the set of vertices in graph C on (acyclic) undirected paths between A and B, except for A and B themselves. (Since the algorithm is continually updating C, $A_C(A,B)$ and $U_C(A,B)$ are constantly changing as the algorithm progresses.)

### PC Algorithm

A.) Form the complete undirected graph C on the vertex set **V**.

B.)

n = 0.

repeat

For each pair of variables A, B adjacent in C, if $A_C(A,B) \cap U_C(A,B)$ has cardinality greater than or equal to n and A, B are d-separated by any subset of $A_C(A,B) \cap U_C(A,B)$ of cardinality n, delete A - B from C.

n = n + 1.

until for each pair of vertices A, B that are adjacent in C, $A_C(A,B) \cap U_C(A,B)$ is of cardinality less than n.

C.) Let F be the graph resulting from step B. For each triple of vertices A, B, C such that the pair A, B and the pair B, C are each adjacent in F but the pair A, C are not adjacent in F, orient A - B - C as A -> B <- C if and only if A and C are not d-separated by any subset of $A_F(A,C) \cap U_F(A,C)$ containing B.

D. repeat

If there is a directed edge A -> B, and an undirected edge B - C, and no edge of either kind connecting A and C, then orient B - C as B -> C. If there is a directed path from A to B, and an undirected edge A - B, orient A - B as A -> B.

until no more arrowheads can be added.

We have run this algorithm on as many as 90 variables in randomly generated sparse graphs. In Monte Carlo simulations on linear models, for large sample sizes (on the order of several thousand) on sparse graphs, the percentage of errors of omission or commission for adjacencies is below 2%. Under the same conditions, the percentage of arrowheads that are erroneously omitted is less than 2%, and the percentage of arrowheads that are erroneously added is about 20%. (See Spirtes, Glymour, and Scheines forthcoming for the details of this Monte Carlo study.)

Following Pearl's terminology, the output of the algorithm when there are no unmeasured common causes is a mixture of directed and undirected edges called a **pattern**. A pattern Π represents a set of directed acyclic graphs. A graph G is in the set of graphs represented by Π if and only if:

1. G has the same adjacency relations as Π.

2. If the edge between A and B is oriented A -> B in Π, then it is oriented A -> B in G.

3. There are no unshielded colliders in G that are not also unshielded colliders in Π.

A **hybrid graph** may contain contain bidirected edges edges. Such graphs may be used to represented the marginal structure on a set of measured variables when unmeasured common causes have edges that collide (in Verma and Pearl's terminology, are "head to head") with edges between measured variables. When there are unmeasured common causes, the output of the PC algorithm can include such bi-directed edges. X - Y denotes that there is an undirected edge between X and Y in a pattern Π, X o-> Y denotes that X and Y are adjacent and there is at least an arrowhead into Y, X -> Y denotes X o-> Y and not Y o-> X, and X <-> Y denotes X o-> Y and Y o-> X. Two vertices are **adjacent** in a hybrid graph if X - Y, X -> Y, or X <-> Y. Two edges **collide** at Y if and only if each edge has an arrowhead directed into Y. The definitions of directed path, undirected path, and d-separability for hybrid graphs are the same as for directed graphs.

Let D be a directed acyclic graph with vertex set U of which the subset O are observable. **The pattern Π of D restricted to O** is the hybrid graph that is the result of applying the PC algorithm to the d-separability relations of D that involve just variables in O. From the marginals over O of distributions faithful to D, the PC algorithm can be used to construct the pattern of D. (The method that Verma and Pearl use to construct the pattern of D is based upon an earlier algorithm described in Spirtes 1990 for constructing graphs when no latent variables are present given faithful input. It is equivalent in output to the PC algorithm, but is too slow to be used on large numbers of variables, and is less reliable in practice because it requires testing high order conditional independence relations.)

The following theorem is a corollary of a theorem proved by Verma and Pearl:

**Theorem 1:** Given a graph G over a set of variables U, a distribution P that satisfies the Markov condition for G, and some subset O of U, for X, Y, and C that are disjoint subsets of O, the Markov condition entails that X is independent of Y conditional on C if and only if X and Y are d-separated by C in the pattern of G over O.

In a pattern Π, a **semi-directed path P from X to Y** is an undirected path from X to Y such that if A occurs before B in P, then in Π the edge from B to A does not have an arrowhead into A. In other words, a semi-directed path can contain both undirected edges and directed edges pointing in the direction from X to Y, but it cannot contain any bidirected edges or edges pointing in the direction from Y to X.

I will show the following;

**Theorem 2:** Let G be a graph over a set of vertices U, and O be a subset of U containing X and Y, and Π the pattern of G over O. If there exists a directed path A from X to Y in G then Π contains a semi-directed path B from X to Y.

Theorem 2 states that if the probability distribution on the set of measured variables is not compatible with a semi-directed path B from X to Y then if that distribution is the marginal of a distribution perfectly represented by a graph on a larger vertex set, the larger graph also contains no directed path from X to Y. In that case, it is possible to know that A is not a cause of B by examining the marginal over the observed variables, even if latent variables are present.

The following theorem states under what conditions it is possible to infer from a pattern the existence of a directed path in a graph G In a directed acyclic graph or a pattern, X , Y, and Z form a **triangle** if and only if X is adjacent to Y and Z, and Z is adjacent to Y.

**Theorem 3:** Let O be a subset of vertices of G containing X and Z, and let the pattern of G for O contain a directed edge X -> Z, no triangle containing X and Z, and a variable C such that C o-> X . Then in G there is a directed path from X to Z.

Furthermore:

**Corollary 1:** Let O be a subset of vertices of G containing X and Z, and let the pattern of G for O contain a vertex C such that C o-> X , and a directed path A from X to Z such that for no adjacent pair U, W on A is there a triangle in Π containing both U and W. Then in G there is a directed path from X to Z.

Theorem 2 and Corollary 1 state conditions under which it is possible to know that A is a (possibly indirect) cause of B simply by examining the marginal over the observed variables, even if latent variables are present.

## 3   Can Theorem 3 Be Strengthened?

Verma and Pearl have recently published a claim (Corollary 1 in Pearl 1991) that entails a stronger version of Theorem 3. A consequence of their Corollary 1, translated into the language of this paper is[1]:

Let O be a subset of vertices of G containing X and Z, and let the pattern Π of G for O contain a directed path P from X to Z, such that for every edge A -> B in P there is a variable C such that C o-> A. Then in G there is a directed path from X to Z.

This claim is false. It is stronger than Theorem 3 because it does not require that each edge in P not be part of a triangle in P order to infer the existence of a directed path from X to Z in G. The graph depicted in Figure 1 (suggested by the proof of Theorem 3) provides a counterexample. In G, there is no directed path from X to Z; however in the pattern of G over the set of variables that does not include T, there is a uni-directed edge from X to Z, and an edge D -> X.

---

[1] The statement of their Corollary 1 is somewhat ambiguous, so it is not immediately obvious that the correct interpretation entails the consequence stated here. However Pearl (personal communication) has confirmed that the interpretation presented here is the intended one.

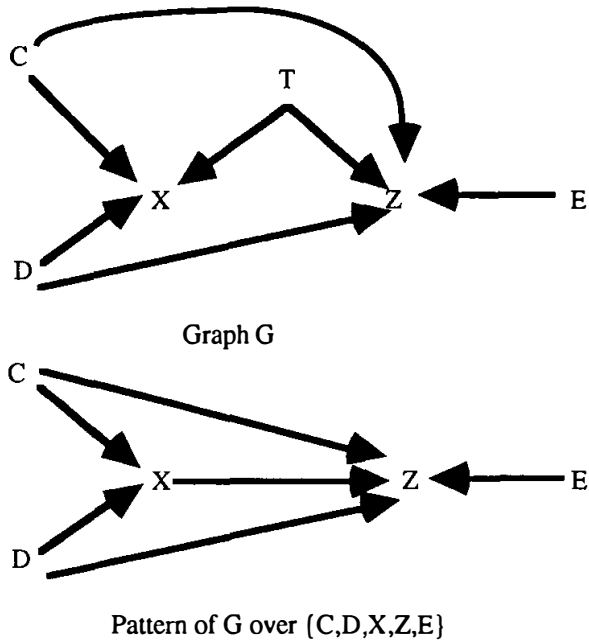

Graph G

Pattern of G over {C,D,X,Z,E}

**Figure 1**

## 4  Appendix

**Lemma 1** (Verma and Pearl): Let O be a subset of U, D a directed acyclic graph over U, and Π the pattern of D restricted to O. Two variables A, B in O are adjacent in Π if and only if there exists a path P between A and B in D satisfying the following two conditions:

(1) every observable node on P (except the endpoints) is a collider along P; and

(2) every collider along P is a shieldable ancestor of either A or B.

where an ancestor S of A is shieldable if and only if every directed path from S to A contains an observable other than A.

**Lemma 2** (Verma and Pearl): For any pattern Π of D over O, A o-> B if and only if there is a node C such that either (1) C is adjacent to B and not A (in Π) and both edges A - B and B - C were induced by paths (of D) which ended pointing at B, or (2) C o-> A in Π and B is a descendent of A in D.

**Lemma 3:** Let G be a directed acyclic graph over a set of vertices U, O be a subset of U containing X and Y, and Π be the pattern of G restricted to O. If there is a directed path $P_1$(X,Y) in G from X to Y that induces an edge between X and Y in Π, then either X -> Y or X - Y in Π.

**Proof.** The proof is a reductio. Assume that there is a directed path P in G from X to Y that induces and edge between X and Y in Π, but neither X -> Y nor X - Y in Π. It follows that Y o-> X in Π. By lemma 1, there is a vertex Z in O such that either Z is adjacent to X and not to Y in Π, and both of the edges between X and Y and X and Z are induced by paths pointing at X, or Z o-> X in Π, and X is a descendant of Y in G. X is not a descendant of Y in G, because Y is a descendant of X in G, and G is acyclic. Suppose then that there is a vertex Z in O such that Z is adjacent to X and not to Y in Π, and both of the edges between X and Y and X and Z are induced by paths pointing at X.

Let P(Z,X) be an undirected path between X and Z that points into X and induces an edge between X and Z in Π, and $P_2$(X,Y) be an undirected path between X and Y that points into X and induces an edge between X and Y in Π. Let R be the first point of intersection of P(X,Z) with $P_2$(X,Y), P(Z,R) be the subpath of P(Z,X) from Z to R, P(R,Y) be the subpath of $P_2$(X,Y) from R to Y, and P(Z,Y) be the concatenation of P(Z,R) and P(R,Y). P(Z,Y) is an undirected path because P(Z,R) and P(R,Y) by construction intersect only at R, and hence P(Z,R) and P(R,Y) contain any given vertex at most once.

Every vertex on P(Z,Y) that is in O is a collider on P(Z,Y). By lemma 1, every vertex in O on P(Z,R) except for Z and R are colliders on P(Z,R), and every vertex in O on $P_2$(R,Y) except for R and Y are colliders on $P_2$(R,Y) (because they are subpaths of paths that induce edges between Z and X, and X and Y respectively.) This shows that each vertex in O with the possible exception of R on P(Z,Y) is a collider on P(Z,Y). I will now show that if R is in O, it is also a collider on P(Z,Y). If R is equal to X, then X is a collider on P(Z,Y) because both P(Z,X) and P(X,Y) are into X. If R is not equal to X, and R is in O, then R is a collider on both P(Z,X) and P(X,Y); hence R is a collider on P(Z,Y). It follows that every vertex on P(Z,Y) that is in O is a collider on P(Z,Y).

Suppose first that every collider on P(Z,Y) is a shieldable ancestor of either Z or Y. By lemma 1, P(Z,Y) induces an edge between Z and Y in Π. It follows from lemma 2 that P(Z,X) and $P_2$(X,Y) do not induce a Y o-> X orientation. This contradicts the assumption.

Suppose next that there is a collider on P(Z,Y) that is not a shieldable ancestor of either Z or Y, and let W be the first such collider after Z. R is the only vertex on P(Z,Y) that may be a collider on P(Z,Y) but not a collider in either P(Z,X) or $P_2$(X,Y). Hence W is either equal to R or a collider on P(Z,X) or $P_2$(X,Y).

In either case Y is a descendant of W. Suppose first that W is a collider on P(Z,X) or $P_2$(X,Y). Because W is not a shieldable ancestor of either Z or Y, by lemma 1, W is a shieldable ancestor of X. X is an ancestor of Y, and W is an ancestor of X, so W is an ancestor of Y. Suppose next that W is equal to R. In this case R is not equal to X, because X is an ancestor of Y and X is in O, and hence X is a shieldable ancestor of Y. Either X is a descendant of R, Y is a descendant of R, or some collider along $P_2(X,Y)$ is a descendant of R. If X is a descendant of R, then Y is a descendant of R, because Y is a descendant of X. If some collider along $P_2(X,Y)$ is a descendant of R, then Y is a descendant of R because each collider on $P_2(X,Y)$ is a shieldable ancestor of either X or Y, and Y is a descendant of X. In any case, then, Y is a descendant of W.

W is an ancestor of Y, but not a shieldable ancestor of Y, so there is a directed path $P(W,Y)$ from W to Y that contains no vertices in O other than Y. Let S be the first point of intersection of $P(Z,Y)$ with $P(W,Y)$, $P(Z,S)$ the subpath of $P(Z,Y)$ from Z to S, $P(S,Y)$ the subpath of $P(W,Y)$ from S to Y, and $P_2(Z,Y)$ the concatenation of $P(Z,S)$ and $P(S,Y)$. S is not a collider on $P_2(Z,Y)$ because the first edge $P(S,Y)$ is not directed into S. Hence every collider on $P(Z,S)$ is a collider on $P_2(Z,W)$. W is the first collider on $P(Z,Y)$ that is not a shieldable ancestor of Z or Y, S either equals W or is before W on $P(Z,Y)$, S is not a collider on $P_2(Z,Y)$, and $P(S,Y)$ contains no colliders; hence every collider on $P_2(Z,Y)$ is a shieldable ancestor of either Z or Y.

Every vertex V on $P_2(Z,Y)$ that is in O is on $P(Z,S)$. V is not equal to S because S is not in O. If V is not equal to S, then V is a collider on $P(Z,S)$, and hence a collider on $P_2(Z,Y)$.

By lemma 1, $P_2(Z,Y)$ induces an edge between Z and Y in P. It follows from lemma 2 that $P(Z,X)$ and $P_2(X,Y)$ do not induce a Y o-> X orientation. This contradicts the assumption.

**Theorem 2:** Let G be a graph over a set of vertices U, and O be a subset of U containing X and Y, and Π the pattern of G over O. If there exists a directed path A from X to Y in G then Π contains a semi-directed path B from X to Y.

**Proof.** Break the path A in G into a series of subpaths such that only the endpoints of the subpaths are in O. Let U be the source and V be the sink of some such arbitrary subpath. There is an edge between U and V in Π by lemma 3. U is prior to V on B. The concatenation of the edges induced by the subpaths are an undirected path B from X to Y in Π. By Lemma 1, it is not the case that V o-> U. By definition of semi-directed path, B is a semi-directed path from X to Y.

**Theorem 3:** Let O be a subset of vertices of G containing X and Z, and let the pattern Π of G for O contain a directed edge X -> Z, no triangle containing X and Z, and a variable D such that D o-> X. Then in G there is a directed path from X to Z.

**Proof.** Since X and Z are adjacent in Π there is a path A in G between X and Z such that every observable node on A is a collider and every collider on A is a shieldable ancestor of X or Z.

If X -> Z in pattern Π arises because of clause (2) of lemma 2, we are done because Z is a descendant of X in G. So suppose X -> Z is oriented by condition (1) of lemma 2. Then there is a path A in G that induces X -> Z and A is into Z in G. If A contains no colliders and is not into X, then A is a directed path from X to Z, and we are done. Otherwise there are two cases: A contains a collider, or A is into X.

First, we consider the case where A is into X. Then there is a path between X and Z that induces an edge in Π, and is into X. By assumption there is a vertex D in Π such that D o-> X. By lemma 2, either there is a vertex C in Π such that C is adjacent to X and not D, and both edges C-X and D-X are induced by paths of G which point at X, or C o-> D in Π and X is a descendant of D in G.

Suppose that the first disjunct is true. In that case, either C is adjacent to Z in Π or it isn't. If it is adjacent to Z, then there is a triangle in Π containing X and Z. If C is not adjacent to Z, then by clause (2) of lemma 2, Z o-> X, contrary to our assumption.

Suppose now that the second disjunct is true. Because X is a descendant of D in G, there is a directed path in G from some variable E in O to X that does not contain any variables in O other than X and E. This path induces an edge between E and X in Π. If E is adjacent to X in Π, then Π contains a triangle containing X and Z; if E is not adjacent to X in Π, then by clause (2) of lemma 2, Z o-> X, contrary to our assumption.

We now consider the case where A is not into X, but there is a collider on A. Let K be the first collider on A after X. Because A is not into X, then there is a directed path from X to K, and hence no directed path from K to X. This implies that K is not an ancestor, and hence not a shieldable ancestor of X. So by clause (2) of lemma 1, K is a shieldable ancestor of Z. The concatenation of the paths from X to K and from K to Z is a directed path from X to Z.

**Corollary 1:** Let O be a subset of vertices of G containing X and Z, and let the pattern Π of G for O contain a vertex C such that C o-> X, and a directed path P from X to Z such that for no adjacent pair U, W on P is there a triangle in Π containing both U and W. Then in G there is a directed path from X to Z.

**Proof.** Since there is a directed path from X to Z in Π, there is a sequence of edges X -> A -> B ... -> Z in Π. By Theorem 3, there is a directed path from X to A in G. Since X -> A in Π, Theorem 3 can next be applied to A -> B, to show that there is a directed path from A to B in G. Repeating this process in turn for each edge on P implies that there is a directed path from X to Z in G.

## Acknowledgements

I wish to thank Clark Glymour for a number of useful discussions and suggestions on the topic of causal inference when latent variables are present. This research was supported in part by a graph with the Office of Naval Research, and the Naval Manpower Research and Development Center under Contract number N00114-89-J-1964. Theorem 3 and a weaker version of Theorem 2 were reported in "Causal Structure among Measured

Variables Preserved with Unmeasured Variables", by Peter Spirtes and Clark Glymour, Laboratory for Computational Linguistics Technical Report No. CMU-LCL-90-5, August, 1990.